\pdfoutput=1
\documentclass[11pt]{article}

\usepackage[final]{acl}
\usepackage{booktabs} 
\usepackage{times}
\usepackage[listings]{tcolorbox}
\usepackage{makecell}
\usepackage{amsmath}
\usepackage{xcolor}
\usepackage{tikz}
\usepackage{float}
\usepackage{tcolorbox}
\tcbuselibrary{skins}
\usepackage{tcolorbox}
\usepackage{setspace}
\usepackage{lipsum}
\usepackage{geometry}
\usepackage{latexsym}
\usepackage{etoc}
\usepackage[utf8]{inputenc}
\usepackage{amsmath} % For \text{} in math mode

\usepackage{tcolorbox}     % for boxed display of algorithm

\usepackage{amsmath} % For math symbols if needed within algorithm
\usepackage{amsfonts} % For math symbols if needed within algorithm
\usepackage[T1]{fontenc}
\usepackage[utf8]{inputenc}

\usepackage{microtype}
\usepackage{booktabs} % Include this in your preamble

\usepackage{inconsolata}

\usepackage{graphicx}
\usepackage{times,latexsym}
\usepackage{url}
\usepackage[T1]{fontenc}

\usepackage{tcolorbox}
\usepackage{algorithm}
\usepackage{algorithmic}
\usepackage{hyperref}
\usepackage{xurl}

\usepackage{epigraph}
\usepackage{enumitem}
\usepackage{algorithmic}
\usepackage{algorithmic}
\usepackage{amsmath}

\usepackage{hyperref}
\usepackage{url}
\usepackage[table]{xcolor}

\usepackage{minitoc}
\usepackage{graphicx}
\usepackage{booktabs}
\usepackage{epigraph}
\usepackage{enumitem}
\usepackage{tcolorbox}
\usepackage{amssymb}

\usepackage{epstopdf}
\usepackage{algorithm}
\usepackage{etoolbox}
\usepackage{tcolorbox}
\usepackage{enumitem}
\usepackage{dblfloatfix}
\title{\textit{From Fragments to Facts:} A Curriculum-Driven DPO Approach for Generating Hindi News Veracity Explanations}

\author{
Pulkit Bansal$^{1}$\thanks{Work done during undergraduate studies at Indian Institute of Technology Patna.} \quad
Raghvendra Kumar$^{2}$ \quad
Shakti Singh$^{3}$ \quad
Adam Jatowt$^{4}$ \enspace
Sriparna Saha$^{2}$ \\
\\
$^{1}$TCS Research, India \\
$^{2}$Department of Computer Science and Engineering, Indian Institute of Technology Patna, India \\
$^{3}$Indian Institute of Technology Patna, India \\
$^{4}$University of Innsbruck, Austria \\
\texttt{pulkitbansal996@gmail.com}\quad\quad\texttt{\{raghvendra\_2221cs27,sriparna\}@iitp.ac.in}
}

\begin{document}
\maketitle

\begingroup
\renewcommand\thefootnote{}
\footnotetext{
\textbf{Project Page:} \href{https://pulkit48.github.io/From-Fragments-to-Facts/}{From Fragments to Facts}
}
\endgroup
\begin{abstract}
  In an era of rampant misinformation, generating reliable news explanations is vital, especially for underrepresented languages like Hindi. Lacking robust automated tools, Hindi faces challenges in scaling misinformation detection. To bridge this gap, we propose \textbf{\textit{DeFactoX}}, a novel framework integrating Direct Preference Optimization (DPO) with Curriculum learning to align machine-generated explanations with human reasoning. Fact-checked explanations from credible sources serve as preferred responses, while LLM outputs highlight system limitations and serve as non-preferred responses. 
  At the core of this framework lies \textbf{\textit{Hin-DPO}}, an enhanced variant of DPO that enriches the loss function with two novel parameters, \textit{Actuality }and \textit{Finesse}, enhancing explanation quality and consistency.
  Experiments with LLMs (Mistral, Llama, Gemma) and PLMs (mBART, mT5) confirm the framework's effectiveness in generating coherent, contextually relevant explanations. 
  % Moreover, owing to the data augmentation capability of our framework, it can be effectively adapted to other low-resource languages, thereby broadening the scope of automated explanation generation for combating misinformation.
\end{abstract}

% \iftaclpubformat
\section{Introduction}

% \epigraph{"Fake news is cheap to produce. Genuine journalism is expensive."}{\textit{Toomas Hendrik Ilves}}

The rise of fake news, fuelled by its low production cost and widespread digital dissemination, poses a significant threat to the integrity of journalism. As Toomas Hendrik Ilves aptly states, ``\textit{Fake news is cheap to produce. Genuine journalism is expensive.}" This disparity is evident in the societal disruptions caused by misinformation, particularly during crises like the COVID-19 pandemic, where unverified claims about treatments and preventive measures fuelled panic and confusion \citep{barua2020effects,roozenbeek2020susceptibility}. Misinformation also exacerbates political polarization, creating deep societal divides \citep{cantarella2023does,bovet2019influence,kumar2024silver,kumar2023multimodal}. Fact-checking platforms face substantial challenges in scaling their efforts, especially in languages like Hindi. Despite over 600 million Hindi speakers\footnote{\url{https://www.britannica.com/topic/languages-by-total-number-of-speakers-2228881}}, automated tools for generating credible, human-like explanations in Hindi remain underdeveloped. Addressing this gap is crucial to supporting fact-checking initiatives and reducing the impact of fake news on society.

% \textbf{Overview:} This paper introduces the \textbf{\textit{DeFactoX}} framework, which evaluates the veracity of Hindi news articles and generates coherent, factually grounded explanations. By analyzing content, \textit{DeFactoX} classifies news as credible or misleading and provides justifications, enhancing transparency in misinformation detection.  

% We construct a synthetic Hindi preference dataset using fact-checked \emph{human explanations as preferred responses} and \emph{LLM outputs as rejected responses}. To improve explanation quality, we enhance Direct Preference Optimization (DPO) \citep{rafailov2024direct} with fact scores, variance reduction, and Curriculum learning. These refinements align model outputs with human reasoning and ensure reliability. Leveraging Curriculum learning for data augmentation, our scalable framework extends to low-resource languages, supporting global misinformation detection and explanation generation.

\textbf{Overview:}  
This paper introduces \textbf{\textit{DeFactoX}}, a framework for Hindi news veracity prediction and explanation generation. The framework first constructs a synthetic preference dataset in which human-written, fact-checked explanations serve as preferred responses and LLM-generated explanations serve as rejected responses, thereby grounding alignment in reliable human reasoning while exposing the model to common erroneous patterns. 

Building on this dataset, we use Curriculum learning  \citep{pattnaik2024enhancing} to progressively train the model to distinguish between explanations of increasing difficulty. To further strengthen alignment, we extend Direct Preference Optimization (DPO) \citep{rafailov2024direct} with a novel loss function, termed \textbf{\textit{Hin-DPO}}, which incorporates two additional parameters: \textbf{\textit{Actuality}}, quantifying the factual correctness of responses, and \textbf{\textit{Finesse}}, measuring hallucination through output instability across generations. Together, these innovations enable \textit{DeFactoX} to generate explanations that are factually accurate, consistent, and trustworthy.

\textbf{Research Gap:} While NLP has made significant strides, most automated explanation generation systems focus on high-resource languages like English and Chinese 
% \cite{wang2020feature,martinez2020information,zhang2020trie,xu2024large,hsu2023prompt}
\citep{wang2020feature,zhang2020trie,xu2024large,hsu2023prompt,kumar2026generation}, leaving Hindi largely underserved. 

Pre-trained LLMs, trained on generalized datasets, struggle to assess the veracity of Hindi news and generate contextually relevant, factually grounded explanations for the veracity predicted. Moreover, fact-checking in Hindi remains predominantly manual, lacking scalable automated solutions. Given Hindi's vast speaker base and the increasing spread of misinformation, it is crucial to develop robust, scalable methods for \emph{automated veracity prediction and explanation generation.} 
% While \emph{our framework is language-agnostic}, we focus on Hindi to address this pressing need and demonstrate its effectiveness in a low-resource setting.

\textbf{Research Questions:}  
This research aims to address the following questions:  

\setlist{nolistsep}
\begin{itemize}[noitemsep]

\item \textbf{RQ-1:} How can automated systems reliably assess the veracity of Hindi news and generate human-like explanations that are coherent and contextually relevant?

\item \textbf{RQ-2:} How effective is incorporating parameters such as \textit{Actuality} (factual accuracy) and \textit{Finesse} (hallucination sensitivity) in aligning model outputs with human preferences?

\item \textbf{RQ-3:} How can Curriculum Learning be combined with DPO to progressively enhance veracity prediction and explanation generation for Hindi news?

% , and what scalable methodologies can extend misinformation detection and explanation generation to other low-resource languages?  

\end{itemize}

\textbf{Research Motivation:} The rapid spread of misinformation in languages like Hindi highlights the need for scalable systems that assess veracity and generate reliable, human-like explanations. Unlike high-resource languages, Hindi lacks robust fact-checking tools, and existing LLMs often struggle with coherence, factual accuracy, and human alignment, motivating the need for novel frameworks that ensure trustworthy explanations.

This work addresses these gaps by refining veracity explanation generation through Direct Preference Optimization (DPO) \citep{rafailov2024direct}, Curriculum learning \citep{pattnaik2024enhancing}, and our enhanced loss function \textbf{\textit{Hin-DPO}}, which integrates the \textbf{\textit{Actuality}} score (inspired by FactScore \citep{min-etal-2023-factscore}) and the \textbf{\textit{Finesse}} score (a variance-based measure of hallucination), ensuring both accuracy and scalability.

\textbf{Contributions:}  
\setlist{nolistsep}
\begin{itemize}[noitemsep]
    \item We create a \emph{synthetic, ranking-based Hindi preference dataset}, where human fact-checked explanations serve as top-ranked responses, and LLM outputs are ranked using an \emph{automated scoring mechanism} that closely aligns with human preferences. 

    \item We propose \textbf{\textit{DeFactoX}}, a two-stage framework that combines curriculum learning with an enhanced preference optimization objective, \textbf{\textit{Hin-DPO}}, which integrates two novel parameters: \textbf{\textit{Actuality}} and \textbf{\textit{Finesse}}.  

    \item To the best of our knowledge, \textbf{\textit{DeFactoX}} is the first framework for automated veracity-driven explanation generation in Hindi. Its data augmentation based design provides a scalable methodology that can facilitate future adaptation to other languages.

\end{itemize}

\begin{figure*}[!htbp]
\centerline{\includegraphics[width=0.90\textwidth]{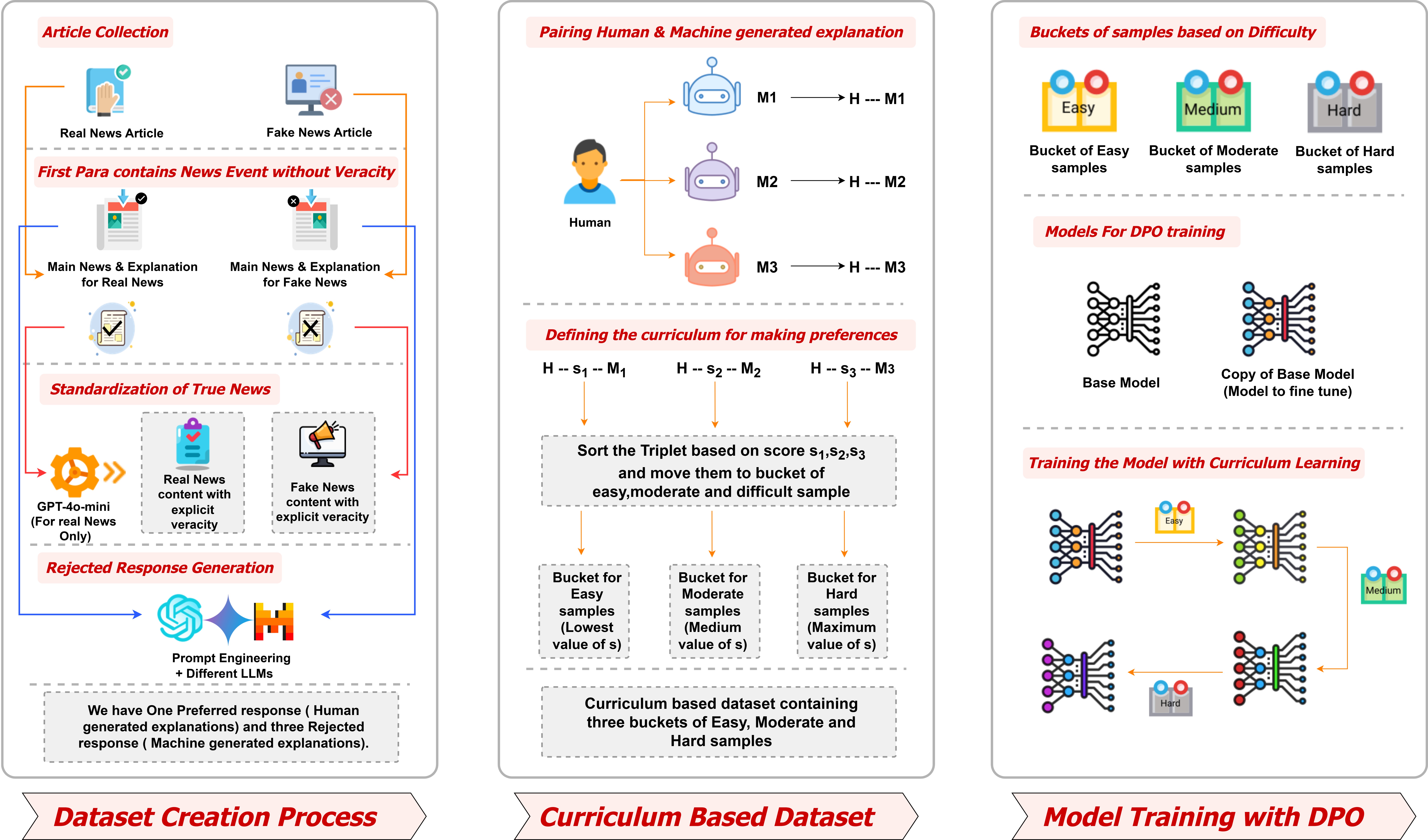}}
\caption{Overview of \textbf{\textit{DeFactoX}} framework. 
\textbf{(Left)} Dataset creation with human-written explanations as preferred responses and LLM-generated explanations as rejected responses.  
\textbf{(Center)} Curriculum-based dataset construction, where samples are ranked and bucketed into easy, moderate, and hard levels.  
\textbf{(Right)} Model training with \textbf{Hin-DPO} under curriculum learning, fine-tuning the model to generate human aligned explanations.}

\label{workflow}
\end{figure*}

\section{Related Works}
% We explore the growth of misinformation detection through veracity explanation alongside advancements in Preference Optimization techniques.

\textbf{Automated Misinformation Detection and Explanation Generation:}  
Recent studies have advanced misinformation detection and explanation generation. \citet{joshi2023explainable} integrated Domain Adversarial Neural Networks (DANN) with LIME to enhance COVID-19 misinformation detection. \citet{chi2022quantitative} proposed QA-AXDS, a scalable, interpretable fake news detection system using dialogue trees. \citet{yao2023end} introduced MOCHEG, a multimodal fact-checking benchmark incorporating textual and visual evidence. \citet{zhou2023synthetic} examined AI-generated misinformation, highlighting linguistic nuances and proposing updated detection guidelines. \citet{gong2024integrating} emphasized socio-contextual cues in ``social explanation” to combat misinformation. \citet{russo2023benchmarking} showed that extractive steps improve abstractive summarization for claim verification. \citet{bilal2024generating} developed a GNN-based rumour verification model leveraging opinion-guided summaries. \citet{yue-etal-2024-evidence} introduced RARG, combining evidence retrieval with RLHF-tuned LLMs, excelling in COVID-19 misinformation detection. The study by \citet{9747214} employed a QA-based framework with attention-driven comparisons for interpretable fact-checking. For a comprehensive review, readers are pointed to the work by \citet{kotonya-toni-2020-explainable}.

\emph{\textbf{Our Novelty:} While previous works primarily target high-resource languages like English and Chinese, our focus is on Hindi, an under-represented language.}

\textbf{Applications and Advancements in Preference Optimization and Curriculum Learning:}  
Recent advancements in preference optimization and Curriculum learning have enhanced model performance across domains. \citet{pattnaik2024enhancing} introduced Curry-DPO, a Curriculum learning-based enhancement of DPO, achieving up to 7.5\% improvement across datasets. \citet{chen-etal-2024-u} proposed a multi-stage Curriculum framework optimizing humour and structure preferences in LLMs. \citet{yin2024self} developed Self-Augmented Preference Optimization (SAPO), surpassing DPO and SPIN across multiple benchmarks. \citet{morimura2024filtered} introduced filtered DPO (fDPO), refining datasets for better training efficiency. \citet{wang2024bpo} proposed Balanced Preference Optimization (BPO), enhancing knowledge depth while maintaining efficiency. \citet{zeng2024token} presented Token-DPO, improving alignment and diversity in LLMs through token-level fine-tuning. \citet{chen2024softmax} proposed Softmax-DPO to enhance recommender systems via user preference modelling. \citet{lai2024step} introduced Step-DPO, improving mathematical reasoning in LLMs with minimal data. \citet{croitoru2025curriculum,kim2024denoising,bansal2026the} have also extended Curriculum learning strategies to diffusion models, demonstrating improved training stability, sample quality, and alignment with human preferences by progressively structuring the learning process. For a comprehensive survey of datasets, theories, variants, and applications in direct preference optimization, readers are referred to \citet{xiao2024comprehensive}.

\emph{\textbf{Our Novelty:} While Direct preference optimization (DPO) has been applied across various domains, our approach is specifically tailored to generating veracity claims and explanations for Hindi news. The research gap in this area highlights challenges faced by Hindi, including data scarcity and the limitations of LLMs trained on multilingual datasets. Our study offers an effective solution to address these challenges.}

\section{Preference Dataset Creation}

To construct our synthetic preference dataset, as shown in Figure \ref{workflow}, we followed a systematic multi-step approach to ensure data quality, uniformity, and relevance to the task. Below, we outline the process in a structured manner.

\subsection{Dataset Selection and Sampling}
Multiple veracity claim misinformation detection datasets \citet{bhardwaj2020hostility,kumar2022fake,sharma2024mmhfnd,bansal2024mmcfnd,10.1145/3493700.3493736,kumar2025sifting} provide Hindi news articles sourced from fact-checking websites, ensuring authentic veracity labels. We selected data instances from \citet{sharma2024mmhfnd}, a comprehensive dataset featuring over 15,000 articles in the fake news category and 13,000+ in the real news category. This dataset was chosen for its extensive coverage of news, spanning from older to recent events, and its sourcing from fact-checking websites, which provide verified veracity labels. To maintain a balanced and manageable dataset, we extracted the \textbf{most recent} 5,000 articles from each class (fake and real). The selection ensures a healthy mix of data samples without being overbearing in size. Further stats are presented in the appendix (Section \ref{dataset-stats}).

\subsection{Characteristics of the Selected Data}  

\textbf{(1)} Each article is sourced from fact-checking websites that not only classify news as fake or real but also provide comprehensive, well-reasoned explanations justifying these classifications. These \textbf{explanations serve as ground truth references} for evaluating model-generated outputs.  

\textbf{(2)} The first paragraph of every article \textbf{strictly contains only the core news content}, intentionally excluding any veracity or reasoning. This neutral presentation ensures that readers, and more importantly, models cannot determine whether the news is fake or real based solely on this segment. 

This design choice is crucial for constructing the preference dataset, as these initial news passages act as inputs for models, which must then generate both veracity predictions (fake or real) and coherent supporting explanations.

\subsection{Observations on Explanations}  

A key observation in the dataset is the \textbf{distinct difference in the writing styles of explanations for true and fake news}.  

\begin{figure}[!htbp]
\centering
\centerline{\includegraphics[width=0.45\textwidth]{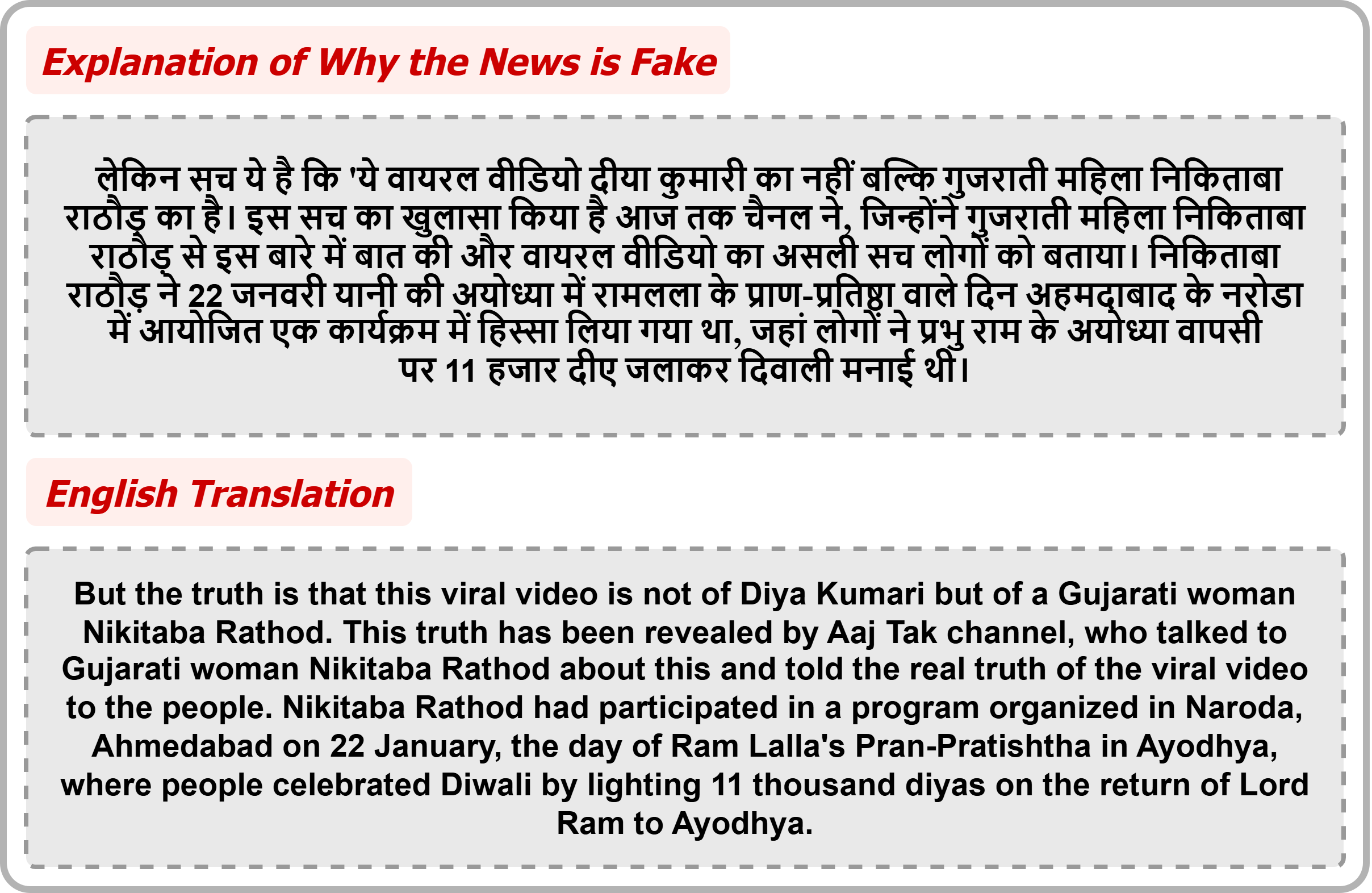}}
\caption{Snippet of fake news explanation with explicit reasoning for its veracity.}
\label{exp}
\end{figure}

\begin{figure}
    \centering
    \includegraphics[width=0.45\textwidth]{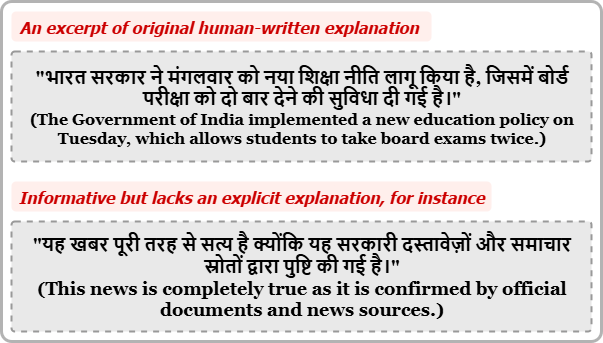}
    \caption{Example of a human-written true news explanation that is primarily informational, summarizing facts without explicitly confirming authenticity.}

    \label{fig:truenews}
\end{figure}

\textbf{(1) True news explanations:} During our manual verification, we observed that many human-written explanations were primarily informational in nature. They tended to summarize the facts of the news story but did not include explicit reasoning or assertive statements confirming its authenticity. Such explanations typically present the news content in a descriptive manner, without additional justification or emphasis on truthfulness as shown in Figure~\ref{fig:truenews}.

\textbf{(2) Fake news explanations:} In contrast, explanations for fake news are far more detailed and explicit. Fact-checkers provide strong declarative statements, rejecting falsehoods and supplementing them with clear justifications, such as evidence-based counterarguments, source verification, and logical reasoning. This explicit mention of veracity is illustrated in Figure \ref{exp}. For a deeper understanding of these explanations, readers may consult the original fact-checking sources: OneIndia\footnote{\url{https://hindi.oneindia.com/fact-check/}}, Vishvas News\footnote{\url{https://www.vishvasnews.com/}}, and Aaj Tak\footnote{\url{https://www.aajtak.in/fact-check}}, which serve as the primary references for this dataset and are certified by the International Fact-Checking Network (IFCN).

\subsection{Standardizing True News Explanations}

While explanations for fake news naturally include explicit reasoning and veracity statements, true news explanations often lack such clarity. Fact-checking sources assume that factual news is self-evident, leading to minimal justifications. This inconsistency makes it difficult for models to learn a uniform veracity-based explanation structure. To address this, we standardized true news explanations by ensuring they explicitly affirm their veracity while preserving factual integrity. This step aligns true news explanations with the structured reasoning seen in fake news explanations. For implementation details and examples, please refer to the appendix (Section \ref{sec:true_news_prompt} \& \ref{true-news-example}).

\subsection{Generating Rejected Responses}

To construct the rejected responses, we used the \textbf{first paragraph of each article, presenting only the core news content,} as input for three state-of-the-art LLMs: \texttt{gpt-4o-mini} \cite{achiam2023gpt}, \texttt{Mistral-7B-v0.1} \cite{jiang2023mistral}, and \texttt{gemini-1.5-flash} \cite{team2024gemini}. These models were selected due to their strong reasoning capabilities and proven performance in NLP tasks \citep{liu-etal-2024-longgenbench,mathur-etal-2024-knowledge,siino2024mcrock,siino2024mistral,trott-riviere-2024-measuring,sato-etal-2024-tmu}.  

Unlike existing approaches that aim to minimize hallucinations in synthetic preference datasets, our goal is fundamentally different: we intentionally generated \textbf{weaker, non-preferred explanations} using the prompt in appendix (Section~\ref{negative_gen_prompt}). We use a simple prompt without additional fine-tuning or safeguards, letting the generated responses naturally reflect the inherent limitations of LLMs.

\textbf{Observations:} Through manual verification of 1,500 sampled rejected responses, we found recurring issues that distinguished them from human-written explanations. Specifically, LLM-generated explanations often:  
\begin{itemize}
    \item Over-emphasized superficial linguistic elements or unusual words in the news text.  
    \item Displayed bias towards stylistic or surface-level framing rather than factual grounding.  
    \item Focused on a single aspect of news instead of covering multiple factual dimensions.

    \item Lacked the depth, balance, and context-awareness consistently present in human-authored explanations.  
\end{itemize}

These weaknesses confirmed their suitability as the negative class for preference optimization. 
% For completeness, the exact prompt used to generate rejected responses is provided in the appendix (Section~\ref{sec:news_prompt}).  

\subsection{Final Preference Dataset Composition}
The final synthetic preference dataset comprises:
\setlist{nolistsep}
\begin{itemize}[noitemsep]
    \item \textbf{Preferred outputs:} Explanations for fake news, sourced from fact-checking websites, and true news explanations, standardized using the prompt described in the appendix 
    % \emph{\textbf{Supplementary}}.
    (Section \ref{sec:true_news_prompt}).
    \item \textbf{Rejected outputs:} Machine-generated explanations, based on the first paragraph of the news articles, produced by three state-of-the-art LLMs: \texttt{gpt-4o-mini}, \texttt{Mistral-7B-v0.1}, and \texttt{gemini-1.5-flash}. 
\end{itemize}
Each data sample contains one positive (preferred) explanation and three negative (rejected) explanations, ensuring a balanced dataset by maintaining a consistent 1:3 ratio. This structure provides equal exposure to both high-quality and suboptimal explanations, helping to improve distinction and generalization. Furthermore, examples of input news and their corresponding non-preferred outputs, are provided in the appendix (Section \ref{example-non-pref-output}).

\section{Methodology}

% Inspired by the work of \citet{pattnaik2024enhancing}, we incorporated curriculum learning to improve model robustness in distinguishing high-quality explanations. Specifically, the three non-preferred responses were ranked according to their alignment with ground truth explanations using a scoring function. The training was organized progressively: the model first learned from the least aligned explanations, then from moderately aligned ones, and finally from the most aligned. This curriculum strategy enabled a gradual transition from easier cases to more challenging ones that require fine-grained distinctions.  

We propose \textit{\textbf{DeFactoX}}, a unified framework for news veracity prediction and explanation generation that emphasizes both factual reliability and explanatory robustness. It builds on two complementary ideas: a curriculum learning strategy that ranks explanations by alignment with ground truth and trains progressively from easier to harder cases, and a domain-aware extension of the standard DPO objective, \textbf{\textit{Hin-DPO}}, which incorporates signals for factual consistency and stability. Together, these components help \textit{DeFactoX} better align with human preferences (see Appendix~\ref{algos}).

\subsection{Explanation Ranking for Curriculum Learning}

To integrate curriculum learning into our framework, we require a mechanism to distinguish explanations by their quality and progressively guide the model from easier to more challenging training cases. We achieve this by scoring non-preferred explanations, ranking them according to their alignment with ground-truth rationales, and then organizing the training sequence based on these ranks. 

\textbf{\emph{Scoring Function for Explanation Ranking:}}  
The scoring function \textit{\textbf{fs}} combines BERTScore \citep{zhang2019bertscore}, ROUGE-L \citep{lin2004rouge}, and METEOR \citep{banerjee2005meteor}.\footnote{The \textit{fs} scoring function used in our experiments is specifically designed and validated for our dataset. For other datasets or tasks, alternative scoring functions may be used.}

\[
\text{\textbf{\textit{fs}}} = \frac{\text{BERTScore} + 3 \times (\text{ROUGE-L} + \text{METEOR})}{4}
\]

\subsection{Validation of the Scoring Function (\textit{\textbf{fs}})}
\label{appendix:fs_validation}

To validate the choice of our scoring function, we conducted an empirical study over 300 randomly sampled explanations (150 True News and 150 Fake News). Each sample was shown to human annotators, who were asked to rank the three rejected responses in order of quality. We then compared three weighting strategies for aligning automatic scores with these human rankings: (i) equal weighting of all metrics (1:1:1), (ii) a moderate weighting giving twice the importance to (ROUGE-L + METEOR) relative to BERTScore (1:2), and (iii) a stronger weighting giving three times the importance to (ROUGE-L + METEOR) relative to BERTScore (1:3).  

The degree of alignment between automatic scores and human-provided rankings was quantified using the Spearman rank correlation \citep{spearman1904proof}, with results summarized in Table~\ref{tab:spearman_alignment}.  

\begin{table}[h]
\centering
\begin{tabular}{l c}
\hline
\textbf{Scoring Strategy} & \textbf{Spearman $\rho$} \\
\hline
1:1 weighted average & 0.63 \\
1:2 weighted average & 0.74 \\
\rowcolor{blue!10}
1:3 weighted average & \textbf{0.81} \\
\hline
\end{tabular}
\caption{Alignment of scoring strategies with human-annotated rankings. 
Here, the ratios indicate the relative weight assigned to BERTScore versus the combined contribution of ROUGE-L and METEOR. 
For example, 1:3 means BERTScore is given weight~1, while the sum of ROUGE-L and METEOR is given weight~3.}
\label{tab:spearman_alignment}
\end{table}

The 1:3 weighting achieved the strongest alignment ($\rho=0.81$) and was therefore adopted in our framework. \emph{Importantly, this scoring mechanism is not proposed as a novel contribution, but rather as a supporting utility to align curriculum-based training with human judgments.} Other weighting schemes or evaluation metrics may be equally suitable in different datasets or applications.

\textbf{Curriculum Learning Strategy:} Explanations were ranked using scoring function \textit{\textbf{fs}} into three levels: \textbf{rank-0} (least aligned with ground truth, having minimum fs), \textbf{rank-1} (moderately aligned), and \textbf{rank-2} (most aligned, maximum fs). This curriculum strategy guided the model from easier to harder distinctions: it was first trained on rank-0 explanations, which are clearly different from the preferred responses and thus easier to identify, followed by rank-1 explanations with moderate similarity, and finally on rank-2 explanations. The rank-2 cases are the most challenging, as they differ only subtly from the preferred responses, requiring the model to capture fine-grained distinctions. This structured progression, akin to human learning, enhances robustness and generalizability.  

\subsection{\textbf{Our \textit{Hin-DPO} Loss function}}
We extend the standard DPO objective by proposing \textbf{\textit{Hin-DPO}}, a modified loss function tailored for news veracity prediction and explanation generation. Hin-DPO incorporates two additional parameters: \textbf{\textit{Actuality}}, which captures the factual correctness of explanations, and \textbf{\textit{Finesse}}, a variance-based measure that quantifies the degree of hallucination. By integrating these factors, \textit{Hin-DPO} refines preference optimization to emphasize both factual accuracy and consistency in explanation generation.

\subsection{\textbf{Explaining Actuality}}
% \label{section:actuality_score} 
\textbf{Rationale:}  
The reliability of AI-generated explanations for news tasks relies heavily on factual accuracy. Given the rise of misinformation and its societal consequences, explanations must align with verifiable facts. To encourage this, we introduce the \textit{Actuality} score as a reward signal, prioritizing factually accurate explanations over those containing incorrect or unverifiable content.

\textbf{Calculation:} 
The \textit{Actuality} score is designed to ensure factual accuracy in news explanations. It leverages GPT-4o-mini’s internal knowledge, with access to a web search tool, to assess factual correctness. The score is computed by extracting key factual statements from a news article, evaluating their correctness, and averaging the results into a single numerical value between 0 and 1. The exact prompting strategy used to compute the \textit{Actuality} score is provided in appendix (Section~\ref{appendix:actuality_prompt}).

\textbf{Justification:}  
Fluent explanations can still be misleading. The \textit{Actuality} score promotes factual consistency by penalizing incorrect explanations and rewarding factually aligned ones. Using an LLM enables detection of subtle inaccuracies, and experimental validation of this score is provided in the appendix (Section~\ref{section:actuality_experiments}).

We clarify that GPT-4o-mini is used for rejected response generation during dataset construction and for sentence-level factual verification during Actuality scoring; however, these usages occur in distinctly different operational settings. Further details are provided in Appendix~\ref{appendix:gpt_usage}.

% \textbf{Justification:}  
% Highly fluent but misleading explanations can appear convincing, leading to misinformation. The \textit{Actuality} score penalizes inconsistent explanations and rewards factual ones. Using an LLM for factuality detection enables reliable identification of subtle inaccuracies. Incorporating this score into training ensures fact-based alignment. Experimental validation of the \textit{Actuality} score are provided in appendix (Section~\ref{section:actuality_experiments}).

\subsection{Explaining \textit{Finesse}}

\textbf{Rationale:} 
In veracity prediction and explanation generation tasks, hallucinations often arise from model uncertainty, which typically manifests as instability across repeated generations for the same input. Empirical evidence from LLMs shows that when a model possesses reliable knowledge, independently sampled responses tend to be consistent, whereas hallucinated content leads to divergent and contradictory generations \citep{manakul2023selfcheckgpt}.

From an uncertainty estimation perspective, variability across stochastic model outputs has long been used as a proxy for predictive uncertainty. Prior work shows that repeated stochastic forward passes can be interpreted as samples from an approximate posterior over model outputs, with variance capturing uncertainty \citep{gal2016dropout}. Similar uncertainty effects have been observed in autoregressive structured prediction, where competing token-level alternatives during sequential decoding lead to variability across generated outputs \citep{malininuncertainty}. More broadly, recent studies and surveys demonstrate that token-level uncertainty signals are effective indicators of factual unreliability and hallucination in large language model outputs \citep{fadeeva2024fact,kang2025uncertainty}. Motivated by these empirical and theoretical insights, we introduce the \textit{Finesse} score to quantify instability in explanation generation, which serves as an indicator of hallucination driven by underlying model uncertainty.

\textbf{Calculation:} 
To compute \textit{Finesse}, we generate five stochastic responses for the same input using a high decoding temperature ($T=0.9$). During decoding, the model produces a probability distribution over the vocabulary at each time step. For each generation run $r$, we aggregate these token-level distributions across the response length $L_r$:
\[
\bar{p}^{(r)} = \frac{1}{L_r} \sum_{t=1}^{L_r} p^{(r)}_{t},
\]
where $\bar{p}^{(r)} \in \mathbb{R}^{|V|}$ represents the overall lexical distribution for the $r$-th generated explanation.

The \textit{Finesse} score is then computed as the average variance across these aggregated distributions over the vocabulary:
\[
\text{\textit{Finesse}} =
\frac{1}{|V|}
\sum_{v \in V}
\mathrm{Var}\big(
\bar{p}^{(1)}(v),
\ldots,
\bar{p}^{(5)}(v)
\big).
\]

This formulation measures the stability of the model’s output probability distributions across multiple independent responses to the same input. Lower \textit{Finesse} values indicate consistent and confident generations, whereas higher values reflect increased variability across responses, signaling greater underlying uncertainty and a higher likelihood of hallucination.

\subsection{Modified DPO Loss Function}

This section presents the modifications made to the DPO loss function by incorporating domain-specific parameters such as \textit{Actuality} and \textit{Finesse} scores.

\begin{table*}[!b]
    \centering
    \scalebox{0.80}
    {
    \begin{tabular}{l|ccccc|ccccc|ccc}
    \toprule
        \textbf{Model $\rightarrow$} & \multicolumn{5}{c}{\textbf{mBART}} & \multicolumn{5}{c}{\textbf{mT5}} & \multicolumn{3}{c}{\textbf{Gemma2-9B}}\\
        
    \midrule
        \textbf{Config$\downarrow$} & \textbf{R-1} & \textbf{R-2} & \textbf{R-L} & \textbf{MT} & \textbf{BS} & \textbf{R-1} & \textbf{R-2} & \textbf{R-L} & \textbf{MT} & \textbf{BS} & \textbf{R-1} & \textbf{R-2} & \textbf{R-L}\\ 
     \midrule
        Base & 13.39 & 6.33 & 9.89 & 18.49 & 70.14 & 15.41 & 7.11 & 10.21 & 19.29 & 70.01 & 29.51 & 18.03 & 22.17 \\ 
        Base+SFT & 14.68 & 8.21 & 10.99 & 20.52 & 71.37 & 16.83 & 8.09 & 11.15 & 20.15 & 72.78 & 30.12 & 18.74 & 23.11 \\ 
        DPO & 16.21 & 9.61 & 12.18 & 20.60 & 73.17 & 17.93 & 9.14 & 13.59 & 22.87 & 73.61 & 30.92 & 19.86 & 25.19 \\ 
        DPO+Act & 17.66 & 10.50 & 11.73 & 22.12 & 75.32 & 19.23 & 9.51 & 14.22 & \textbf{24.67} & 76.35 & 32.68 & 20.55 & 27.13 \\ 
        DPO+Fin & 17.96 & 10.95 & \textbf{13.73} & 22.63 & 76.09 & 19.81 & \textbf{9.53} & 15.97 & 24.11 & 75.33 & 33.41 & 21.26 & 26.59 \\
        \rowcolor{blue!10}
        Hin-DPO & \textbf{18.98} & \textbf{11.85} & 13.19 & \textbf{23.50} & \textbf{77.19} & \textbf{20.29} & 9.45 & \textbf{16.89} & 24.39 & \textbf{77.32} & \textbf{33.55} & \textbf{21.64} & \textbf{27.91} \\ 
       $\Delta$ (vs DPO) & +2.77 & +2.24 & +1.01 & +2.90 & +4.02 & +2.36 & +0.31 & +3.30 & +1.52 & +3.71 & +2.63 & +1.78 & +2.72\\

    \midrule
    \textbf{Model $\rightarrow$} & \multicolumn{5}{c}{\textbf{Mistral-7B}} & \multicolumn{5}{c}{\textbf{Llama3.1-8B}}  & \multicolumn{3}{c}{\textbf{Gemma2-9B}} \\ \midrule
        \textbf{Config$\downarrow$} & \textbf{R-1} & \textbf{R-2} & \textbf{R-L} & \textbf{MT} & \textbf{BS} & \textbf{R-1} & \textbf{R-2} & \textbf{R-L} & \textbf{MT} & \textbf{BS} & \textbf{MT} & \multicolumn{2}{c}{\textbf{BS}} \\ 
     \midrule
        Base & 26.21 & 16.69 & 22.72 & 26.11 & 76.34 & 32.12 & 19.91 & 25.02 & 30.61 & 77.52 & 29.12 & \multicolumn{2}{c}{76.88} \\ 
        Base+SFT & 26.89 & 18.02 & 24.36 & 27.15 & 78.27 & 33.46 & 21.03 & 27.45 & 32.97 & 79.82 & 30.62 & \multicolumn{2}{c}{78.45} \\ 
        DPO & 28.04 & 19.77 & 25.76 & 28.92 & 79.34 & 34.56 & 22.00 & 28.73 & 34.53 & 80.98 & 31.21 & \multicolumn{2}{c}{79.59}  \\ 
        DPO+Act & 29.65 & 20.76 & 27.13 & \textbf{30.18} & 81.94 & 35.71 & 23.12 & 29.23 & 36.13 & 82.42 & 33.10 & \multicolumn{2}{c}{82.23} \\ 
        DPO+Fin & 29.79 & 21.12 & 27.34 & 29.51 & 81.55 & 36.51 & 22.33 & 30.19 & 36.23 & 83.79 & 33.17 & \multicolumn{2}{c}{82.52} \\ 
        \rowcolor{blue!10}
        Hin-DPO & \textbf{30.82} & \textbf{22.07} & \textbf{28.13} & 29.87 & \textbf{82.95} & \textbf{37.13} & \textbf{24.07} & \textbf{31.22} & \textbf{37.25} & \textbf{84.73} & \textbf{33.84} & \multicolumn{2}{c}{\textbf{83.67}} \\ 
        $\Delta$ (vs DPO) & +2.78 & +2.30 & +2.37 & +0.95 & +3.61 & +2.57 & +2.07 & +2.49 & +2.72 & +3.75 & +2.63 & \multicolumn{2}{c}{+4.08}\\

    \bottomrule
    \end{tabular}
    }
    \caption{Performance comparison across models. \textbf{Abbreviations:} R-1: ROUGE-1, R-2: ROUGE-2, R-L: ROUGE-L, MT: METEOR, BS: BERTScore, Act: Actuality, Fin: Finesse. Bold values denote best performance. The blue-shaded row corresponds to \textbf{\textit{Hin-DPO}}. The $\Delta$ rows indicate relative improvement of \textit{Hin-DPO }over DPO.}

    \label{tab:results}
\end{table*}

\begin{equation*}
\text{Let } r_w = \frac{\pi_\theta(y_w \mid x)}{\pi_{\text{ref}}(y_w \mid x)}, \quad
r_l = \frac{\pi_\theta(y_l \mid x)}{\pi_{\text{ref}}(y_l \mid x)}
\end{equation*}
\begin{equation*}
\begin{split}
S(x, y_w, y_l) &= \frac{1}{v + \epsilon} \bigg[ (1+s_w) \log r_w \\
&\quad - \max(0.01, s_l) \log r_l \bigg]
\end{split}
\label{eq:score}
\end{equation*}
Then the \textit{Hin-DPO} loss function is defined as:
\begin{equation*}
L_{\text{Hin-DPO}}(\pi_\theta; \pi_{\text{ref}}) = - \mathbb{E}_{(x, y_w, y_l) \sim D} \Big[ 
\log \sigma \big( \beta \cdot S \big) \Big]
\label{eq:dpo_loss_1}
\end{equation*}

\noindent
Here $\pi_\theta(y \mid x)$ and $\pi_{\text{ref}}(y \mid x)$ denote the probabilities assigned by the learned policy and the reference policy, respectively, for a response $y$ given input $x$. The dataset sample $(x, y_w, y_l) \sim D$ consists of an input $x$, a preferred response $y_w$, and a rejected response $y_l$. The \textit{Actuality} scores $s_w$ and $s_l$ quantify the factual accuracy of the preferred and rejected responses, respectively. The hyperparameter $\beta$ controls the preference weighting, while $v$ represents the \textit{Finesse} score. $\epsilon$ is a learnable parameter. The gradient analysis of \textit{Hin-DPO} is represented in appendix (Section~\ref{grad-analysis}).

\textbf{Utilization of \textit{Actuality}:}  
We leverage \textit{Actuality} scores to modulate the weighting of preferred and rejected responses. Specifically, $(1 + s_w)$ amplifies the log probability of the preferred response, ensuring that factually accurate explanations are prioritized. Conversely, $\max(0.01, s_l)$ is applied to the rejected response, penalizing factually weak explanations without over-penalization.

\textbf{Impact of \textit{Finesse}:}  
The \textit{Finesse} score measures explanation uncertainty and is integrated into the Hin-DPO loss as a scaling factor. Explanations with low variance, indicating greater stability and consistency, receive higher preference weight, while unstable or potentially hallucinatory explanations with high variance are down-weighted. When variance approaches one, the loss reduces to standard DPO behavior. This mechanism encourages the model to favor factually reliable and consistent explanations during training.

\begin{table*}[!ht]
\centering
\scalebox{0.80}
{
\begin{tabular}{l|ccccc|ccccc}
\hline
\textbf{Model $\rightarrow$} & \multicolumn{5}{c}{\textbf{mT5}} & \multicolumn{5}{c}{\textbf{LLaMA3.1-8B}} \\
\midrule
\textbf{Config$\downarrow$} & R-1 & R-2 & R-L & MT & BS & R-1 & R-2 & R-L & MT & BS \\
\midrule
DPO (w/o CL) & 17.93 & 9.14 & 13.59 & 22.87 & 73.61 & 34.56 & 22.00 & 28.73 & 34.53 & 80.98 \\
DPO (with CL) & 18.62 & 9.33 & 15.37 & 22.96 & 75.02 & 35.07 & 22.85 & 30.00 & 35.01 & 82.04 \\
Hin-DPO (w/o CL) & 18.78 & 9.26 & 15.22 & 23.50 & 76.19 & 35.14 & 22.17 & 29.74 & 35.22 & 82.01 \\
\rowcolor{blue!10}
Hin-DPO (with CL) & 20.29 & 9.45 & 16.89 & 24.39 & 77.32 & 37.13 & 24.07 & 31.22 & 37.25 & 84.73 \\
\bottomrule
\end{tabular}
}
\caption{Ablation study on mT5 and LLaMA3.1-8B models with and without Curriculum Learning. Curriculum Learning consistently improves performance. \textbf{Abbreviations:} R-1: ROUGE-1, R-2: ROUGE-2, R-L: ROUGE-L, MT: METEOR, BS: BERTScore. The blue-shaded row corresponds to our proposed \textbf{\textit{Hin-DPO}} method.}
\label{tab:ablation_combined}
\end{table*}

\section{Experimental Setup}
We fine-tuned five models, comprising three Large Language Models (LLMs): Gemma-2-9B-It \citep{gemma_2024}, Llama-3.1-8B-Instruct \citep{dubey2024llama}, and Mistral-7B-Instruct-v0.3 \citep{jiang2023mistral} and two Pre-trained Language Models (PLMs): mBART-large-50 \citep{tang2020multilingual} and mT5-large \citep{xue-etal-2021-mt5}. The quality of the generated explanations was assessed using three key metrics: BERTSCORE \citep{zhang2019bertscore}, ROUGE-1,2, L score \citep{lin2004rouge} and METEOR score \citep{banerjee2005meteor}. Given the involvement of Hindi, we utilized the Polyglot tokenizer \citep{al-rfou-etal-2013-polyglot} to evaluate ROUGE-1, 2, L, and METEOR scores. Hyperparameters are presented in the appendix (Section \ref{hyperparameters}).

\section{Results and Analysis}
In this section, we present the experimental results and analyze the effectiveness of the proposed framework, \textbf{\textit{DeFactoX}}. Table~\ref{tab:results} summarizes the performance of different training strategies across automatic evaluation metrics, illustrating the impact of preference-based optimization and task-specific signals on explanation generation quality. Veracity prediction results are reported separately in the appendix (Section~\ref{section:veracity_prediction_results}).

Across different model backbones, incorporating DPO-based training improves performance over the Base+SFT baseline. Variants such as \textit{DPO}, \textit{DPO+Actuality}, and \textit{DPO+Finesse} show steady gains in all metrics, reflecting improvements in semantic alignment and explanation quality. Among all methods, \textit{Hin-DPO} achieves the strongest results, indicating that combining task-specific preference signals with curriculum sequencing benefits explanation generation in the Hindi news domain. These findings address \textbf{RQ1} by showing that preference-based optimization improves explanation quality, and \textbf{RQ2} by underscoring the importance of language and task-aware adaptation.

The improvements are consistent across different model architectures, including Gemma2-9B and Llama3.1-8B, suggesting that the observed trends are not specific to a single backbone. In particular, higher BERTScore values indicate better semantic similarity to reference explanations, while gains in ROUGE and METEOR reflect improved lexical overlap and fluency. Together, these results suggest that DPO-based fine-tuning enhances both surface-level and semantic aspects of generated explanations. The inclusion of curriculum learning further contributes to these gains, as it allows models to progressively adapt to increasingly aligned explanation preferences, addressing \textbf{RQ3}.

\begin{table}[!htbp]
\centering
\begin{tabular}{lcc}
\toprule
\textbf{Method} & \textbf{Gemma2-9B} & \textbf{Llama3.1-8B} \\
\midrule
Base+SFT & 3.29 & 3.07 \\
DPO & 3.92 & 3.87 \\
\rowcolor{blue!10}
Hin-DPO & \textbf{4.12} & \textbf{4.23} \\

\bottomrule

\end{tabular}
\caption{Human Evaluation Scores (0–5) for Gemma2-9B and Llama3.1-8B. Bold indicates best performance. The blue-shaded row corresponds to our method.}
\label{tab:humaneval}
\end{table}

\textbf{Human Evaluation:}  
We conducted a human evaluation on 800 explanations (400 real and 400 fake), assessed by three student evaluators. Each explanation was rated on a 0–5 scale using predefined criteria, with scores averaged across annotators. Inter-annotator agreement was measured using Spearman correlation, yielding a coefficient of 0.71, indicating substantial and consistent agreement. The detailed evaluation criteria are provided in the appendix (Section~\ref{appendix:human_eval_criteria}). As shown in Table~\ref{tab:humaneval}, both \textbf{Llama3.1-8B} and \textbf{Gemma2-9B} achieve higher human evaluation scores under \textit{Hin-DPO} compared to other training strategies, aligning with trends observed in automatic metrics and suggesting improved explanation quality.

\textbf{Ablation Study:}  
We further analyze the contribution of curriculum learning by comparing models trained with and without it. Table~\ref{tab:ablation_combined} shows that curriculum learning provides consistent improvements across ROUGE, METEOR, and BERTScore for mT5, and more pronounced gains for Llama3.1-8B. These results indicate that gradually introducing preference-aligned training signals helps stabilize learning and improves explanation quality. The individual effects of the \textit{Actuality} and \textit{Finesse} components are reported in Table~\ref{tab:results}, showing that each contributes positively to overall performance.

\section{Conclusion}  
In this work, we presented \textbf{\textit{DeFactoX}}, a framework for veracity-focused explanation generation for Hindi news. At its core, \textit{Hin-DPO} extends preference optimization by incorporating \textit{Actuality} to ensure factual correctness and \textit{Finesse} to reduce hallucination, while Curriculum Learning progressively aligns model outputs with human reasoning. 
\textit{DeFactoX} offers practical utility: media houses and fact-checkers can use it to detect misleading narratives, while end-users benefit from contextually accurate explanations that strengthen their ability to distinguish truth from misinformation. Despite these advances, challenges remain, including limited availability of high-quality fact-checked data and difficulty handling highly complex or domain-specific claims. Future directions include extending \textit{DeFactoX} to other low-resource languages through multilingual transfer and incorporating human-in-the-loop feedback to further enhance explanation quality.

\section*{Limitations}

\textit{DeFactoX} has several limitations. The availability of high-quality, fact-checked Hindi data remains limited, which constrains the diversity and domain coverage of the training and evaluation sets and may reduce effectiveness for highly specialized or technical news requiring substantial background knowledge. While the framework is evaluated on Hindi, its applicability to other low-resource languages requires additional validation, resources, and access to native speakers.
Other than that the evaluation is restricted to models under 10B parameters due to computational constraints, and comparisons with larger reasoning-oriented models are not included. In addition, the Actuality score relies on external LLMs, which may introduce biases or inaccuracies, and the computation of Finesse requires multiple generations per input, leading to increased computational cost.

% \section*{Limitations} 

% \textit{DeFactoX} faces several limitations. High-quality, fact-checked Hindi datasets are limited, restricting the diversity and domain coverage of training data. The framework may struggle with highly specialized or technical news where background knowledge is crucial. While the approach is designed for Hindi, its adaptation to other low-resource languages requires further validation. Additionally, the Actuality score relies on external LLMs, which may themselves contain biases or inaccuracies, and Finesse computation requires multiple output generations, increasing computational costs.

\section*{Ethics Statement} 

\textit{DeFactoX} is designed to support responsible misinformation mitigation by generating reliable, human-aligned explanations for underrepresented languages like Hindi. Human evaluators were involved to ensure alignment with human judgment and reduce automated bias. Users are advised to exercise caution and verify high-stakes claims, particularly in politically or medically sensitive contexts. While Actuality and Finesse reduce hallucination and factual errors, source data biases may persist. All data used is publicly available fact-checked news, and no private or sensitive user information was included.

\bibliography{custom}
\clearpage

\appendix
\section{Appendix}
\etocsettocstyle{}{}
\etocsetnexttocdepth{subsection}

\localtableofcontents

\subsection{Major Hyperparameters}\label{hyperparameters}

This section details the resources and configurations used in our experiments. Dataset generation with \texttt{Mistral-7B-v0.1} was performed on an NVIDIA RTX 3090 GPU (24 GB), whereas \texttt{gpt-4o-mini} and \texttt{gemini-1.5-flash} relied on smaller GPUs, as their API-based tasks required minimal computational resources. Fine-tuning of the LLMs was carried out on an L40S GPU with 45 GB of memory to ensure efficient processing. The dataset was split into 75\% training, 5\% validation, and 20\% testing subsets. For data generation, \texttt{gpt-4o-mini} and \texttt{Mistral-7B-v0.1} employed a temperature of 0.7, top\_k of 50, and top\_p of 0.95, while \texttt{gemini-1.5-flash} used default parameters. Direct Preference Optimization (DPO) alignment was performed over 10 epochs with a learning rate of $1\times10^{-4}$, a batch size of 2, and a beta value of 0.6, taking approximately 48 hours, with fine-tuning conducted in parallel.

\subsection{Dataset Statistics}\label{dataset-stats}
In this section, we provide a more comprehensive overview of the dataset, covering label distribution and model-specific explanation statistics (Tables \ref{tab:token} \& \ref{tab:count}). This structured overview captures the source and average length of explanations, offering transparency into how explanations were curated and generated. It also helps readers understand the composition of training samples used in preference optimization.

\begin{table}[t]
\centering
\small
\begin{tabular}{lccc}
\toprule
\textbf{Source} & \textbf{Type} & \textbf{Count} & \textbf{Avg. Tokens} \\
\midrule
Human & Preferred & 10K & 124.7 \\
GPT-4o-mini & Non-Preferred & 10K & 110.4 \\
Gemini-1.5 & Non-Preferred & 10K & 112.0 \\
Mistral-7B & Non-Preferred & 10K & 98.1 \\
\bottomrule
\end{tabular}
\caption{Comparison of explanation sources and statistics.}
\label{tab:token}
\end{table}

\begin{table}[htbp]
\centering
\begin{tabular}{l c}
\toprule
\textbf{Label Type} & \textbf{Count} \\
\midrule
Fake News Samples & 5000 \\
Real News Samples & 5000 \\
\midrule
\textbf{Total} & \textbf{10000} \\
\bottomrule
\end{tabular}
\caption{Distribution of Fake and Real news samples in the dataset.}
\label{tab:count}
\end{table}

\subsection{Prompt for Standardizing True News Explanations}
\label{sec:true_news_prompt}

To ensure uniformity in true news explanations, we employed prompt engineering with the \textbf{GPT-4o-mini model} \citep{achiam2023gpt}. The goal was to make true news explanations explicitly state their veracity, aligning them with the structured reasoning found in fake news explanations.

\begin{tcolorbox}[colback=blue!5!white, colframe=blue!75!black, title=\emph{Prompt for Explicit True News Reasoning}]
I will provide you with an article containing a verified news story along with related contextual information.  
Your task is to rewrite the article as an explanation in Hindi, explicitly emphasizing that the news is true. Follow these rules strictly:  

\begin{enumerate}[noitemsep]
    \item Use only the information provided in the article. Do not add, remove, or fabricate any content.  
    \item Insert at least two explicit affirmations stating that the news is true, and repeat this emphasis naturally within the explanation.  
    \item Preserve the original paragraphing and logical sequence of the article, without introducing unnecessary structural changes.  
    \item Ensure the tone is factual, clear, and objective, avoiding exaggeration or speculation.  
    \item The final output must be entirely in Hindi.  
\end{enumerate}  
\end{tcolorbox}

This prompt ensures:
\begin{enumerate}
    \item \textbf{Consistency:} True news explanations explicitly affirm their veracity, aligning them with fake news explanations.
    \item \textbf{Factual Integrity:} The process avoids introducing biases while preserving the original content.
    \item \textbf{Linguistic Uniformity:} All explanations are generated in Hindi for consistency.
\end{enumerate}

\subsection{Prompt for Generating Non-Preferred Explanations}
\label{negative_gen_prompt}
The following prompt was used to generate weaker, non-preferred explanations for comparison with preferred model outputs. It was intentionally kept simple, without additional fine-tuning or safeguards, to capture the natural shortcomings of LLMs.

\begin{tcolorbox}[colback=red!5!white, colframe=red!75!black, title=Prompt for News Explanation]
\emph{Task: I will provide you with a news article. Your task is to do the following:\newline
1. Predict whether the news is fake or real.\newline
2. Provide a detailed explanation for your prediction.\newline
Ensure your response is written as a flowing paragraph, avoiding bullet points, numbering, or any other structured format. The explanation should naturally justify your prediction, without adding any extraneous information.\newline 
Here is the news article: \texttt{article}\newline  
Answer in the form of a detailed paragraph.
}\label{p2}
\end{tcolorbox}

\subsection{Prompt for \textit{Actuality} Score}
\label{appendix:actuality_prompt}

To compute the \textit{Actuality} score, we use a structured prompt that instructs the LLM to extract salient factual statements from a news article and evaluate their factual correctness. The final score is obtained by averaging binary factuality labels assigned to the extracted statements.

\subsection{Example of Standardizing True News Explanations}\label{true-news-example}
To illustrate our standardized prompting approach, Figure \ref{exp-a2} presents an example with the input (a verified Hindi news article on PM Narendra Modi's 2024 election speech) and the corresponding standardized explanation generated using GPT-4o-mini.

\begin{small}
\begin{tcolorbox}[colback=green!5!white, colframe=green!75!black, title=Prompt for \textit{Actuality} Score]
\emph{\textbf{Task:} You will be given a news article. Follow these steps:
\setlist{nolistsep}
\begin{enumerate}[noitemsep]
    \item Extract up to 15 of the most important and factually relevant sentences from the article.
    \item For each extracted sentence, assess its factual correctness:
    \begin{itemize}[noitemsep]
        \item Label each sentence as \textbf{1} if it is factually accurate.
        \item Label it as \textbf{0} if it contains factual errors.
    \end{itemize}
    \item Compute the \textbf{average} of all the labels (1s and 0s).
\end{enumerate}
\textbf{Output:} Return only the factual consistency score as a single numerical value (e.g., 0.75). Do not include any additional explanations, calculations, or extracted sentences.
\textbf{Here is the news article:} \{article\}
\textbf{Answer:}
}
\end{tcolorbox}
\end{small}

\textbf{Input (Original Article):}  
The article reports PM Modi's rally in Andhra Pradesh with Chandrababu Naidu and Pawan Kalyan, highlighting NDA’s unity, the ``400+ seats'' slogan, and themes of progress and development. While factually accurate, the input provides no explicit reasoning about its veracity.

\textbf{Output (Standardized Explanation):}  
The standardized explanation explicitly affirms the truthfulness of the news (e.g., ``This news is completely true''), while retaining all factual details, restructuring content for clarity, and maintaining strict Hindi language consistency. Strategic repetition of verification statements reinforces authenticity without altering the original content.

\subsection{Experimental Validation of \textit{Actuality} Score:}
\label{section:actuality_experiments}
Since the \textit{Actuality} score relies directly on the factual predictions of GPT model, its reliability is strongly tied to the efficacy of this model. To evaluate this, we conducted a human study comparing GPT’s predictions with independent human judgments, as shown in Table \ref{tab:actuality_labels} \& \ref{tab:actuality_confusion}. A total of 200 explanations were randomly sampled, with 100 from true news and 100 from fake news. From each explanation, two sentences were extracted, yielding 400 factual claims. Each claim was fact-checked by three independent student evaluators, who had access to Google Search and ChatGPT with browsing tools, and assigned binary labels as factual (1) or non-factual (0). These labels were then compared against GPT-4o-mini’s binary predictions used in the computation of the \textit{Actuality} score.

\begin{table}[H]
\centering
\begin{tabular}{lccc}
\hline
\textbf{Label Type} & \textbf{(1)} & \textbf{(0)} & \textbf{Total} \\
\hline
Human Labels & 230 & 170 & 400 \\
GPT Predictions & 260 & 140 & 400 \\
\hline
\end{tabular}
\caption{Distribution of human labels and GPT-4o-mini predictions across 400 factual claims. `1' represents Correct and `0' represents Incorrect.}
\label{tab:actuality_labels}
\end{table}

\begin{table}[H]
\centering
\begin{tabular}{lc}
\hline
\textbf{Comparison} & \textbf{Count} \\
\hline
True Positives (TP) & 205 \\
False Positives (FP) & 55 \\
True Negatives (TN) & 115 \\
False Negatives (FN) & 25 \\
\hline
\end{tabular}
\caption{Confusion matrix comparison between GPT-4o-mini predictions and human judgments.}
\label{tab:actuality_confusion}
\end{table}

Overall, GPT-4o-mini achieved an accuracy of 80.0\%, precision of 78.8\%, recall of 89.1\%, and F1-score of 83.7\%. These results demonstrate strong alignment with human fact-checking and validate the use of GPT-4o-mini as the underlying model for computing the \textit{Actuality} score. Nevertheless in principle, any capable language model could be employed to generate the factuality assessments on which the score depends.  

\subsection{Clarification on GPT-4o-mini Usage}
\label{appendix:gpt_usage}

We clarify that GPT-4o-mini is used for rejected response generation during dataset construction and for sentence-level factual verification during Actuality scoring; however, these usages operate under distinct settings.

The two usages differ in purpose, granularity, and operational setup, as summarized in Table~\ref{tab:gpt_usage_table}. Rejected response generation is performed at the document level without external retrieval, whereas Actuality evaluation operates at the sentence level with retrieval support. This separation ensures that the model does not assess its outputs within the same generative setting, thereby mitigating potential methodological concerns.

\begin{table*}[htbp]
\centering
\small
\begin{tabular}{lcc}
\toprule
\textbf{Aspect} & \textbf{Rejected Response Generation} & \textbf{Actuality Evaluation} \\
\midrule
Role & Generate full explanation & Verify factual correctness \\
Input & Full news article & Individual explanation sentences \\
Granularity & Document-level generation & Sentence-level evaluation \\
Output & Free-form explanation text & Veracity judgments \\
External Retrieval & Not used & Web search enabled \\
\bottomrule
\end{tabular}
\caption{Distinction in GPT-4o-mini usage across different stages of the framework.}
\label{tab:gpt_usage_table}
\end{table*}

\subsection{Veracity Prediction Performance}
\label{section:veracity_prediction_results}
In addition to explanation generation, the proposed framework jointly outputs veracity labels, allowing direct evaluation of misinformation detection performance. Table~\ref{tab:veracity_results} reports Accuracy and F1-score across two backbone models and training strategies. Results show consistent improvements from supervised fine-tuning (SFT) to DPO, with the proposed Hin-DPO achieving the best performance for both Gemma2-9B and Llama3.1-8B. Notably, Hin-DPO attains 80.6\% accuracy and 78.4\% F1-score on Gemma2-9B, and 81.2\% accuracy and 78.9\% F1-score on Llama3.1-8B. These results demonstrate that the framework can reliably detect misinformation while simultaneously generating veracity-aligned explanations. Although models optimized solely for classification may achieve higher standalone performance, our approach offers a complementary advantage by integrating accurate veracity prediction with transparent and interpretable reasoning.

\begin{table}[H]
\centering
\begin{tabular}{lcc}
\hline
\textbf{Method} & \textbf{Gemma2-9B} & \textbf{Llama3.1-8B} \\
\hline
Base + SFT & 74.1 / 71.8 & 72.9 / 70.4 \\
DPO & 77.8 / 75.2 & 76.4 / 73.9 \\
Hin-DPO & \textbf{80.6 / 78.4} & \textbf{81.2 / 78.9} \\
\hline
\end{tabular}
\caption{Veracity prediction performance across backbone models and training strategies. Values are reported as Accuracy / F1-score.}
\label{tab:veracity_results}
\end{table}

\subsection{Human Evaluation Criteria}
\label{appendix:human_eval_criteria}

To complement automatic metrics, we conducted a human evaluation using ground-truth explanations provided by professional fact-checking sources. Annotators were instructed to compare each model-generated explanation against the corresponding human-written explanation and evaluate quality along four clearly defined and non-overlapping dimensions. All criteria were rated on a discrete scale from 0 to 5, where higher scores indicate better quality.

\begin{itemize}[noitemsep]
    \item \textbf{Factual Accuracy:} Measures whether the explanation is factually correct and free from hallucinations when compared to the ground-truth explanation. A score of 5 indicates complete factual consistency, while lower scores reflect partial inaccuracies or unsupported claims.

    \item \textbf{Rationale Alignment:} Evaluates how well the explanation captures the core reasoning, evidence, and veracity justification present in the ground-truth explanation. High scores indicate strong alignment with the underlying rationale, whereas low scores indicate missing, distorted, or irrelevant justification.

    \item \textbf{Explanatory Completeness:} Assesses whether the explanation sufficiently covers all critical aspects necessary to justify the veracity decision. Explanations that address the claim comprehensively and without major omissions receive higher scores.

    \item \textbf{Expression Quality:} Measures the clarity, coherence, and readability of the explanation. High scores are assigned to explanations that are well-structured, logically organized, and easy to understand, while low scores indicate confusing, disorganized, or poorly articulated text.
\end{itemize}

Each explanation was independently evaluated by three student annotators with prior experience in news verification tasks. Annotators were blind to model identity. Final scores were computed by averaging ratings across annotators and across the four evaluation dimensions.

\subsection{Example of Non-Preferred Outputs}\label{example-non-pref-output}

To illustrate the composition of the synthetic preference dataset, Figure \ref{exp-a3} presents an input article about Rahul Gandhi’s viral speech video, which critics claimed distorted Mahatma Gandhi’s philosophy. The controversy stemmed from a selectively cropped clip that misrepresented his statement, drawing sharp political reactions.

\textbf{Non-Preferred Outputs (Machine-Generated Explanations)}  
The bottom section of Figure \ref{exp-a3} shows explanations generated by GPT-4o-mini, Mistral-7B-v0.1, and Gemini-1.5-Flash. Their key shortcomings include:

% \textbf{GPT-4o-mini:} Asserts authenticity but ignores the core issue, which is the cropped video, while lacking fact-checking references and contextual analysis.  

% \textbf{Mistral-7B-v0.1:} Misclassifies the article as ``fake'' based on writing style, offering no factual reasoning or engagement with the video’s distortion.  

% \textbf{Gemini-1.5-Flash:} Provides a balanced response but fails to emphasize the edited nature of the video or cite concrete verification methods.  

\textbf{Model 1: GPT-4o-mini} \\
The explanation is considered non-preferred due to its lack of precision and depth. While it correctly asserts that the news is authentic, it fails to address the core misinformation issue: the selective editing of Rahul Gandhi’s speech. A preferred explanation would note that the viral video was intentionally cropped to exclude his immediate correction, which distorted the intended meaning. Moreover, the response does not cite fact-checking sources or verifiable evidence, instead relying on vague statements such as ``typical for political figures.” It also misses a critical analysis of how the manipulated video spread through political reactions. Although it mentions reactions from leaders, it does not contextualize their role in amplifying the misinformation. A stronger explanation would explicitly identify the manipulation, reference fact-checking timelines, and clarify Gandhi’s corrected statement, ensuring a clear, fact-based debunking of the viral claim.  

\textbf{Model 2: Mistral-7B-v0.1} \\
This explanation lacks relevance and critical engagement with the article’s content. It dismisses the piece as ``fake” based purely on writing style, arguing that its ``simple and clear” or ``formal and professional” tone indicates spuriousness. These are not valid indicators of news credibility, as professional reporting often uses such tones. The explanation further fails to address the actual issue: the misleading nature of the viral video and the misrepresentation of Gandhi’s words. A more appropriate response would highlight the intentional cropping of his speech and its distortion. Additionally, the model does not engage with fact-checking resources or logical evidence, instead relying on unsupported stylistic judgments. This lack of reasoning makes the explanation inadequate for identifying and debunking misinformation.  

\textbf{Model 3: Gemini-1.5-Flash} \\
This model produces a more balanced explanation, but it still suffers from key reasoning gaps, making it non-preferred. While it acknowledges the political event and Gandhi’s misstatement, it overlooks the critical detail that the viral video was edited to distort his words. The response treats Gandhi’s correction as proof of authenticity, but misses the fact that the correction was intentionally omitted in the circulated clip. The explanation also briefly mentions ``verifying through reliable sources,” which is positive, but it does not specify how verification should be done or reference fact-checking outlets. A stronger explanation would emphasize the role of editing, highlight the political motivations behind the distortion, and recommend consulting fact-checking platforms. Such an approach would provide a more precise, context-aware, and fact-based debunking.

Overall, these explanations lack precision, factual grounding, and critical reasoning, underscoring why they are deemed non-preferred.

\begin{figure*}[!htbp]
\centerline{\includegraphics[width=0.95\textwidth]{explanations66.jpg}}
\caption{Snippet of True news transformation.}
\label{exp-a2}
\end{figure*}

\begin{figure*}[!htbp]
\centerline{\includegraphics[width=0.90\textwidth]{explanations77.jpg}}
\caption{Snippet of Non-Preferred Response Generation.}
\label{exp-a3}
\end{figure*}

\subsection{Gradient Analysis of Hin-DPO}\label{grad-analysis}

\subsubsection{Objective Function}

The Hin-DPO objective is given by:
\begin{equation}
L_{\text{Hin-DPO}}(\pi_\theta; \pi_{\text{ref}}) 
= -\mathbb{E}_{(x,y_w,y_l)\sim\mathcal{D}} \big[\log \sigma(u)\big],
\end{equation}
where
\begin{equation}
u = \frac{\beta}{v+\epsilon}\Big((1+s_w)r_w - \max(0.01,s_l)\,r_l\Big).
\end{equation}

where we define the log-ratio terms as
\begin{equation}
r_w = \log \frac{\pi_\theta(y_w \mid x)}{\pi_{\text{ref}}(y_w \mid x)}, 
\qquad 
r_l = \log \frac{\pi_\theta(y_l \mid x)}{\pi_{\text{ref}}(y_l \mid x)}.
\end{equation}

\subsubsection{Loss Function Formulation}

\begin{equation}
r(x,y) = \beta \log \frac{\pi_\theta(y\mid x)}{\pi_{\text{ref}}(y\mid x)} + \beta \log Z(x).
\end{equation}

The reward models for our context, associated with the preferred and rejected responses, respectively, are expressed as:
\begin{align}
r_w(x,y) &= (1+s_w)\, r(x,y), \\
r_l(x,y) &= \max(0.01,s_l)\, r(x,y).
\end{align}

The preference probability is
\begin{align}
P(y_w \succ y_l \mid x) 
&= \frac{\exp(r_w(x,y_w))}{\exp(r_w(x,y_w)) + \exp(r_l(x,y_l))} \\
&= \frac{1}{1+\exp\!\big(r_l(x,y_l)-r_w(x,y_w)\big)} \\
&= \sigma \!\Big(\beta \big[(1+s_w) r_w - \max(0.01,s_l) r_l\big]\Big).
\end{align}

\subsubsection{Gradient Derivation}

Let the sigmoid argument be
\begin{equation}
u = \frac{\beta \big((1+s_w) r_w - \max(0.01,s_l)\, r_l\big)}{v+\epsilon}.
\end{equation}

The gradient of the objective is
\begin{equation}
\nabla_\theta L_{\text{Hin-DPO}} =
-\mathbb{E}_{(x,y_w,y_l)\sim\mathcal{D}}
\big[(1-\sigma(u))\, \nabla_\theta u \big].
\end{equation}

We compute
\begin{equation}
\resizebox{0.95\linewidth}{!}{$
\nabla_\theta u =
\frac{\beta}{v+\epsilon}\Big[(1+s_w)\nabla_\theta r_w - \max(0.01,s_l)\nabla_\theta r_l\Big].
$}
\end{equation}

Substituting back gives
\begin{equation}
\nabla_\theta L_{\text{Hin-DPO}} =
-\mathbb{E}_{(x,y_w,y_l)\sim\mathcal{D}}
\left[
\frac{\beta^2 \sigma(u)}{v+\epsilon}\, C
\right],
\end{equation}
where
\begin{equation}
C = (1+s_w)\nabla_\theta r_w - \max(0.01,s_l)\,\nabla_\theta r_l.
\end{equation}

\subsection{Algorithms for Dataset Creation and Hin-DPO Training}\label{algos}

\begin{algorithm*}[htbp]
\caption{Dataset Creation from Fact-Checked News Articles}
\label{alg:dataset-creation-algo}
\begin{algorithmic}[1]
\REQUIRE Scraped news article \( A \) containing main news \( N \) and explanation \( E \), Large Language Models \( \{LLM_1, LLM_2, LLM_3\} \), Scoring function \( S \)
\ENSURE Dataset \( D \) with explanations categorized by quality

\STATE \textbf{Step 1: Segregate Explanation and Main News}
\STATE Extract main news \( N \) and ground-truth explanation \( E_{GT} \) from article \( A \)

\STATE \textbf{Step 2: Generate LLM Explanations}
\FOR{each model \( LLM_i \) in \( \{LLM_1, LLM_2, LLM_3\} \)}
    \STATE Provide \( N \) as input to \( LLM_i \) and obtain predicted explanation \( E_i \)
\ENDFOR

\STATE \textbf{Step 3: Compute Scores for Explanations}
\FOR{each predicted explanation \( E_i \)}
    \STATE Compute score \( S(E_i) \) using scoring function \( S \)
\ENDFOR

% \newpage
\STATE \textbf{Step 4: Rank Explanations}
\STATE Sort explanations \( \{E_1, E_2, E_3\} \) in ascending order based on \( S(E_i) \)

\STATE \textbf{Step 5: Bucketize Explanations}
\STATE Define three score-based categories:
\begin{itemize}
    \item \textbf{Low-quality bucket} \( B_L \) $\leftarrow$ Explanations with lowest scores
    \item \textbf{Medium-quality bucket} \( B_M \) $\leftarrow$ Explanations with mid-range scores
    \item \textbf{High-quality bucket} \( B_H \) $\leftarrow$ Explanations with highest scores
\end{itemize}

\STATE \textbf{Step 6: Construct Final Dataset}
\STATE Form dataset \( D \) by concatenating explanations in the order:
\[
D = B_L \cup B_M \cup B_H
\]

\RETURN \( D \)
\end{algorithmic}
\end{algorithm*}

\begin{algorithm*}[htbp]
\caption{Our Hin-DPO Training Algorithm}
\label{alg:dpo-training}
\begin{algorithmic}[1]
\REQUIRE Training dataset ad Dataloader \( D \) with win and lose samples (paired or unpaired), initial model parameters \( \theta_0 \), reference model \( \pi_{\text{ref}} \), number of iterations \( T \), scaling factor \( \beta \), temperature parameter \( \tau \)
\STATE Initialize model \( \pi_{\theta} \) with parameters \( \theta_0 \)
\STATE Set \( \pi_{\theta} \) to training mode and \( \pi_{\text{ref}} \) to evaluation mode
\FOR{iteration = 1 to \( T \)}
    \FOR{each batch in \( D \)}
        \STATE Initialize running mean \( \mu \gets 0 \) and running variance \( \sigma^2 \gets 0 \) \COMMENT{Running statistics for probability distribution}
        \STATE Set \( \text{num\_iter} = 5 \) \COMMENT{Number of iterations for variance computation}
        \FOR{iter = 1 to \( \text{num\_iter} \)}
            \STATE Compute logits for the preferred response:
            \STATE \( \text{logits} \gets \pi_{\theta}(\text{pref\_ids}, \text{pref\_mask}).\text{logits} \)
            \STATE Compute probabilities:
            \STATE \( \text{probs} \gets \exp(\text{log\_probs(logits, pref\_ids)}) \)
            \STATE Update Mean and Variance:
            \STATE \( \mu \gets \mu + \frac{\text{probs} - \mu}{\text{iter}} \)
            \STATE \( \sigma^2 \gets \sigma^2 + (\text{probs} - \mu) \times (\text{probs} - \mu) \)
        \ENDFOR
        \STATE Compute final variance:
        \STATE \( \sigma^2 \gets \frac{\sigma^2}{\text{num\_iter} - 1} \)
        \STATE Get log probabilities for preferred and dispreferred responses using \( \pi_{\theta} \):
        \STATE \( \text{model\_pref\_log} \gets \text{log\_prob}(\pi_{\theta}(\text{pref\_ids}, \text{pref\_mask}), \text{pref\_ids}) \)
        \STATE \( \text{model\_dispref\_log} \gets \text{log\_prob}(\pi_{\theta}(\text{dispref\_ids}, \text{dispref\_mask}), \text{dispref\_ids}) \)
        \STATE Get log probabilities for preferred and dispreferred responses using reference model \( \pi_{\text{ref}} \):
        \STATE \( \text{ref\_pref\_log} \gets \text{log\_prob}(\pi_{\text{ref}}(\text{pref\_ids}, \text{pref\_mask}), \text{pref\_ids}) \)
        \STATE \( \text{ref\_dispref\_log} \gets \text{log\_prob}(\pi_{\text{ref}}(\text{dispref\_ids}, \text{dispref\_mask}), \text{dispref\_ids}) \)
        \STATE Compute Hin-DPO loss:
        \STATE {\small \( \text{loss} \gets \text{Hin-DPO\_loss}(\text{model\_pref\_log}, \text{model\_dispref\_log}, \text{ref\_pref\_log}, \text{ref\_dispref\_log}, \sigma^2, \beta) \)}
        \STATE Backpropagate loss:
        \STATE \( \text{loss.backward()} \)
        \STATE Update model parameters:
        \STATE \( \theta \gets \text{optimizer.step()} \)
    \ENDFOR
\ENDFOR
\end{algorithmic}
\end{algorithm*}

\end{document}